\documentclass[letterpaper]{IEEEtran}
\usepackage{geometry}

\IEEEoverridecommandlockouts
\usepackage{cite}
\usepackage{amsmath,amssymb,amsfonts}
\usepackage{graphicx}
\usepackage{textcomp}
\usepackage{xcolor}
\usepackage{cite}
\usepackage{amsmath,amssymb,amsfonts}
\usepackage{graphicx}
\usepackage{textcomp}

\usepackage[noend]{algpseudocode}
\usepackage{algorithm}

\usepackage{caption}
\graphicspath{{images/}} 
\usepackage{makeidx}         
\usepackage{multirow}
\usepackage{mathtools}
\usepackage{amssymb}
\usepackage{tabu}
\usepackage{booktabs}
\usepackage[export]{adjustbox}

\usepackage{caption}
\usepackage{graphicx}
\usepackage{makeidx}         
\usepackage{graphicx}        
\usepackage{pgfplots}
\pgfplotsset{compat=1.13}
\usepackage{booktabs}
\usepackage{amsmath}
\usepackage{csvsimple}
\usepackage{pgfplots}
\usepackage{multicol}
\def\BibTeX{{\rm B\kern-.05em{\sc i\kern-.025em b}\kern-.08em
    T\kern-.1667em\lower.7ex\hbox{E}\kern-.125emX}}
\begin{document}

\title{An Evolving Population Approach to Data-Stream Classification with Extreme Verification Latency}

\author{\IEEEauthorblockN{Conor Fahy and Shengxiang Yang}
\IEEEauthorblockA{Institute of Artificial Intelligence, School of Computer Science and Informatics\\
De Montfort University, Leicester LE1 9BH, UK\\
Email: conor.fahy@dmu.ac.uk, syang@dmu.ac.uk}}

\maketitle

\begin{abstract}
Recognising and reacting to change in non-stationary data-streams is a challenging task. The majority of research in this area assumes that the true class label of incoming points are available, either at each time step or intermittently with some latency. In the worse case this latency approaches infinity and we can assume that no labels are available beyond the initial training set. When change is expected and no further training labels are provided the challenge of maintaining a high classification accuracy is very great. The challenge is to propagate the original training information through several  timesteps, possibly indefinitely, while adapting to underlying change in the data-stream. In this paper we conduct an initial study into the effectiveness of using an evolving, population-based approach as the mechanism for adapting to change. An ensemble of one-class-classifiers is maintained for each class. Each classifier is considered as an agent in the sub-population and is subject to selection pressure to find interesting areas of the feature space. This selection pressure forces the ensemble to adapt to the underlying change in the data-stream.
\end{abstract}

\begin{IEEEkeywords}
Data Stream Analysis,  Scarcity of Labels , Extreme Verification Latency , Evolutionary Algorithms
\end{IEEEkeywords}

\section{Introduction}
A data-stream is a continuously arriving sequence of data. Let $S=[(x_i, y_i)^t]_{t=0}^\infty$ represent a stream where $x_i$ is a point which describes a class $y_i$, at time $t$. Often the underlying operation which generates $S$ is non-stationary and distributions within $S$ will drift as $S$ progresses. Classification models which are static and not adaptive to these changes are likely to degrade over time and their predictions will become increasingly less reliable. In the worst case, change in the stream can be so severe that our non-adaptive model becomes entirely obsolete.

This change can happen in a number of ways. Concept drift occurs when the characteristics of the incoming data points ($x_i$) change, or the relationship between the data ($x_i$) and the target class ($y_i$) changes. Formally, we describe this relationship as a conditional probability $P^t(y_i|x_i)$ where vector $x_i$ describes class $y_i$ at time $t$ . Concept drift can be described as being real or virtual. In virtual concept drift there is a change in $P(x_i)$ which does not result in a change in $P(y_i|x_i)$. Here, we could say that the underlying distribution, or features of $x_i$ have changed but that $x_i$ still describes a class $y_i$. When real drift occurs $P(y_i|x_i)$ changes. This change might happen gradually, maybe as virtual drift initially, or the change could be sudden without any change in $P(x_i)$. When real drift occurs, $P^t(y_i|x_i) \neq P^{t+ \delta}(y_i|x_i)$  after some time $\delta$.

In a fully supervised environment, the true class label $y_i$ of $x_i$, is available and there is no \emph{scarcity of labels}, however classification in non-stationary settings when there is a scarcity of labels represents a greater challenge but is often a practical reality when processing data streams. Here, the true class labels of stream samples are not available immediately and there is a \textit{verification latency}. 

Latency $L$ is in the range $[0$,  $\infty]$  and three categories of latency are outlined in \cite{latency2}: null latency when $L = 0$. Intermediate latency where $0 < L < \infty$, and \textit{extreme} latency where $L \to \infty$, which is the worst case where the true label is never available.

This study focuses on the third, and worst, case of verification latency.  We assume that some label information is provided in an initialisation phase but subsequently never again. Furthermore, some form of change is expected and the classification model must adapt to this change without access to labels. 
Such a setting is referred to as \textit{Initially Labelled Non-stationary Streaming} (ILNS) \cite{ditzler} or \textit{Extreme Verification Latency} (EVL) \cite{compose}. For the remainder of this paper we adopt the latter, $EVL$.

This study evaluates population-based approaches for updating and maintaining the classification model. In the proposed method the classification model is a set of one-class-classifier ensembles. Each ensemble is considered to be a sub-population and each individual classifier within the ensemble is an agent within the population. Each agent is subject to selection pressure which forces the population to evolve and adapt to the changing data distribution. We experiment with two bio-inspired approaches, namely: Genetic Algorithm (GA), and Particle Swarm Optimisation (PSO).

The remainder of this paper is outlined as follows: A review of the existing EVL literature is provided in Section \ref{sec:RW}. Our proposed method is outlined in Section \ref{sec:Method}. The experimental section is outlined in Section \ref{sec:Exp}. This includes the data used, metrics, results, peer-comparisons, and a complexity analysis. Finally, Section \ref{sec:conc} provides conclusions and future work.

\section{Related Work}
\label{sec:RW}
An approach to address the problem of EVL involves representing data as a combination of parametric distributions. This concept was initially introduced in the Arbitrary Sub Populations Tracker (APT) \cite{APT}. APT operates based on three assumptions: 1) the drift is gradual and systematic, 2) all sub-populations exist in the training data, and no new sub-populations emerge later in the data stream, and 3) the drift remains constant. However, APT is computationally expensive and relies on assumptions that may not always be practical. The assumption of constant drift affecting each class identically is often unrealistic. Nevertheless, the assumption of gradual drift is typically necessary and prevalent in most EVL algorithms.

When assuming gradual drift, a drifting distribution will exhibit overlap between successive windows. In such cases, the central region of the distribution, known as the \textit{core}, will have the most substantial overlap with the distribution in subsequent time steps. Core Support Extraction (CSE) is a popular strategy for handling EVL. CSE involves identifying instances within the core region of each distribution at time $t$, referred to as Core samples (CS), while discarding non-core samples.

COMpacted POlytope Sample Extraction (COMPOSE) \cite{compose} is an example of an EVL approach employing CSE. COMPOSE employs the $\alpha$-shape \cite{alphaShape}, a generalization of the convex hull, to identify the core support region for each class. However, the computational complexity associated with calculating the $\alpha$-shape makes COMPOSE impractical for high-dimensional data streams. To address this, Learning Extreme Verification Latency with Importance Weighting (LEVELIW) \cite{levelIW} proposed an importance weighting scheme as a substitute for the $\alpha$-shape. Both methods achieve similar results, but LEVELIW offers improved speed. Another proposal to replace the $\alpha$-shape is discussed in \cite{composeGMM}.

FAST COMPOSE \cite{fastcompose} is a novel method that eliminates Core Support Extraction entirely. The idea behind FAST COMPOSE is to propagate all labeled points from one window to the next and transfer the labels from previously labeled points to unknown points in the latest window.

AMANDA \cite{amanda} presents another CSE approach. After a training phase, incoming unlabeled data is classified using a Semi-Supervised Learning (SSL) classifier. Samples in each class are then evaluated using Kernel Density Estimation (KDE) and subsequently filtered based on their KDE scores. Only the densest samples from each class are propagated to the next time step.

SLAYER (Semi-Supervised Stream Learning with density based drift) \cite{slayer} proposes an alternative to CSE by utilizing clustering. A similar approach is outlined in Stream Classification Algorithm Guided by Clustering \cite{scarg} 

A micro-cluster is an $n$-dimensional sphere characterized by its center, $c$, and radius, $r$. It was initially proposed as a density clustering mechanism \cite{onDemandStream} and has been widely employed in the stream clustering literature \cite{cedas, COCEL, flockstream}. The supervised counterpart of a micro-cluster is called a micro-classifier (MClassification), which also records a class label \cite{Mclassif}. The label of an incoming point $p$ is predicted by its nearest micro-cluster $m\phi_i$ and MClassification updates $m\phi_i$ with $p$. 
Bio-Inspired approaches to data stream analysis have been explored in FlockStream \cite{flockstream} (inspired by the flocking behaviour of birds) and there have been approaches based on the observed behaviour of ants \cite{acsc, ants2}.

A survey on EVL methods is given in \cite{EVLsurvey} and an overview of classification with scarcity of labels, which includes but not limited to the EVL setting, is given in \cite{SoLsurvey}. 

\section{Proposed Method}
\label{sec:Method}
An incoming Stream $S$ is processed in tumbling windows of size $w$ (in a tumbling window, a fixed size, non-overlapping chunk of data is considered). Each incoming point $p_i$ in $w$ is passed to the set of ensembles ($e_1, \dots, e_n \in E$) and a prediction $\hat{y}$ is returned. At the end of each window, the fitness of each base-classifier is calculated and the population of classifiers is updated to adapt to the underlying stream.

\subsection{Classification}
\label{sec:Classification}
The proposed method consists of a set of ensembles $E = \{e_1, \dots, e_n\}$ where $n$ is the number of classes present in the training data and $e_i = \{\phi_1, \dots, \phi_j\}$ where $j$ is a user supplied parameter which determines the ensemble size and $\phi$ is a base-classifier. In the following, we use micro-classifiers as the base one-class-classifier (though any distance based or density based one-class classifier could be used). A micro-classifier is an $n$-dimensional sphere, represented as $[{c, r, y, f}]$, where $c$ is the centre of the micro-classifier in $D$ dimensions, $r$ is the radius of the micro-classifier, $y$ is the associated label, and $f$ is the classifier's current fitness. A point $p$ is recognised by a classifier if it falls inside the classifier's boundary; otherwise, the point is unrecognised by this component of the ensemble. We estimate the confidence of the prediction by using the distance from the centre $c$ to point $p$, formally:
\begin{equation}
\begin{split}
dist({\phi_i}_c, p) \leq {\phi_i}_r =\\
\begin{cases}
1: prediction~{\phi_i}_y, & confidence: dist({\phi_i}_c, p)\\
0: unrecognised, & confidence: dist({\phi_i}_c, p)
\end{cases}
\label{eq:classif}
\end{split}
\end{equation}

Point $p$ is passed to each member of the ensemble and the most confident classifier's prediction is returned. If $p$ falls outside the recognised area of the feature space, then the point is classified as `unrecognised'. In these instances, $p_i$ might represent noise, outlier or could be an indication of change in the stream. In some cases, it might be more appropriate to make no prediction if a point is unrecognised, in other instances it might be better to predict every point. In the latter, the ensemble can return the prediction of the classifier most similar to $p_i$. This process is outlined in Algorithm \ref{fig:classifyAlg}.

\begin{algorithm}[!thb]
\caption{Classification}
\label{fig:classifyAlg}

$\boldmath{Input:}$ Ensemble E, point $p$ \\
$\boldmath{Output:}$ Prediction $\hat{y}$ 
\begin{algorithmic}[1]
\State classifiersWhichRecognisePoint $\leftarrow \{\}$
\State LabelsSet $\leftarrow \{\}$
\For {$e_i$ in $E$}
    \For {$\phi_i$ in $e_i$} // for each classifier in ensemble
        \State $d \leftarrow$ distance($\phi_{i_{centre}}, p$)
	    \If{$(d \leq \phi_{i_{rad}})$}  // if point recognised
	      \State $\phi_{i_{dist}} \leftarrow d$
	      \State $\phi_{i_{fitness}} \leftarrow \phi_{i_{fitness}} + 1$
	      \State classifiersWhichRecognisePoint $\leftarrow$ $\phi_i$
            \If{$\phi_y$ not in LabelsSet}
                \State $LabelsSet$ $\leftarrow$ $\phi_y$
            \EndIf
        \EndIf
    \EndFor
\EndFor
\State $C \leftarrow \arg\min_{dist}$(classifiersWhichRecognisePoint)  
\State $\hat{y} \leftarrow C_{y}$
\If{($|LabelsSet| > 1$)} //  there is conflict
    \For {($\phi_i$ in classifiersWhichRecognisePoint)}
        \State {$\phi_{i_{fitness}} \leftarrow 0$}    
    \EndFor
\EndIf
\State \Return $\hat{y}$
\end{algorithmic}
\end{algorithm}

\subsection{Initialising the Ensemble}
\label{sec:Init}
Given a training set containing $n$ classes where $y_1, \dots, y_n \in Y$, the trained classifier $E$ will be composed of $n$ ensembles, each ensemble $e_i$ is a one class classifier specialised in one area of the feature space.
To initialise $E$ two user supplied variables are required; the number of classifiers in each ensemble ($numAgents$) and the radius of each base classifier (the sensitivity of these values will be discussed in Section \ref{sec:CA}). First, the training data is split by class and an ensemble is created for each class. Training samples in each class $y_i \in Y$ are added sequentially. The first instance $x_i$ creates the first micro-classifier with c =  $x_i$, r = radius, y = $y_i$, and $f$ = 0. Subsequent points $x_i$ are passed to existing micro-classifiers to check if they are recognised. If so, the training point is ignored. However, if the point is unrecognised by any of the existing classifiers, it forms a new classifier. This process continues while the number of classifiers is less than or equal to $numAgents$. If this limit is reached before all training samples have been considered, it is typically a sign that the value for $\phi_r$ is too small. Conversely, if training is complete and $|e_i| < numAgents$, we generate additional $k$ classifiers ($k = numAgents - |e_i|$) based on the distribution of existing classifiers. We sample (without replacement) an existing classifier's centre $\phi_{rand{_c}}$ and alter the value of a random dimension using a Gaussian distribution, a new classifier is created from this point and added to $e_i$. 

\subsection{Calculating the fitness function}
\label{sec:Fitness}
If we had access to the ground truth, we could compare each classifier's prediction with the correct label and use this measure as the classifier's fitness. However, in the EVL setting, we do not have access to the underlying truth.

In the proposed method an assumption is made that change in the stream is gradual. $P(x)$ will drift, potentially leading to a change in $P(y|x)$, but $P(y|x)$ will not change without a change in $P(x)$. In the gradual-change assumption, interesting areas of the feature space in one window will have the most overlap with interesting areas in subsequent windows. Once the ensemble has been initialised with labelled data, these interesting areas have been identified and subsequently a $local$ exploration and adaption to the gradual underlying change is required.

Classifiers which recognise incoming points are deemed fit and classifiers which are no longer recognising points (or fewer points) are deemed unfit. This will happen when the underlying distribution has drifted from this part of the feature space. When a classifier makes a prediction (even if this prediction is not used by the ensemble), its fitness increases and after a window the fitness of each classifier decreases linearly ($\phi_f(t +1) = \phi_f(t) - 1$)), so classifiers which we once deemed to be fit gradually become less fit and have fewer opportunities to reproduce or continue to subsequent populations. Alternately, classifiers which are recognising a lot of points are interpreted as being in an interesting part of the feature space and so have a greater chance to reproduce and explore their locality.  
To prevent overlap and situations where classifiers $A$ and $B$ with associated labels $y_a$ and $y_b$ both recognise a point and return contradicting predictions, we penalise the classifiers which make conflicting predictions. These classifiers have a fitness of zero which promotes separation. 

We evaluate two population-based approaches to update the ensemble: A swarm-based approach in Particle Swarm Optimsation (PSO), and an evolutionary approach with a Genetic Algorithm (GA). PSO is an optimisation technique which draws inspiration from the flocking behaviour observed in nature. Each classifier is considered a `particle' and these particles move through the feature space following the fittest members of the swarm and biasing their movement towards historically interesting areas of the space. The position of a particle is influenced by its own best past location and the best past location of the swarm. In the proposed method, the number of swarms is equal to the number of classes in the initialisation phase, so each class is described by a swarm consisting of $n$ particles, where each particle is a classifier.

The GA is inspired by population genetics and belongs to the family of evolutionary computation methods. In the proposed method, each class in the initialisation phase is considered a separate sub-population consisting of $n$ agents. The fittest members of each sub-population are selected (proportional to their fitness) as `parents'. In this phase only the centre of the classifier is considered (not the radius or label). Selected parent solutions are recombined, mutated and the `children' solutions are passed to the next generation. This recombination and mutation constitute a local search and solutions from sub-population $A$ can not combine with solutions from sub-population $B$. The aim is that fitter classifiers will have more of an opportunity to explore their
\begin{table}[!thb]
\centering		
\caption[Training Data Used in COCEL Experiments]{Training Data Used in Experiments}
\small
\label{tab:data}
\begin{tabular}{llccc}\toprule
		& Dataset	& Classes  & Features & Samples \\
		\midrule
		&   $1CDT$      & 2 & 2 & 16,000   \\
		&	$4CR$       & 4 & 2  & 144,000 \\
		&	$1CSurr$    & 2 & 2  & 55,283  \\
		&	$GEARS$     & 2 & 2  & 200,000 \\
		&	$UG-2C-2D$  & 2 & 2  & 100,000 \\
		&	$UG-2C-3D$  & 2 & 3  & 200,000 \\
		&	$UG-2C-2D$  & 2 & 5  & 200,000 \\
		&	$4CRE-V1$   & 4 & 2  & 125,000 \\
		&	$4CRE-V2$   & 4 & 2  & 183,000 \\
		&   $KeyStroke$ & 4 & 10 & 1,600   \\
		\bottomrule	
\end{tabular}
\end{table}
local space and can adapt to gradual change in this way. 

Parent solutions are selected using 2-way tournament selection. Two candidates are randomly selected from the population and a copy of the fittest is passed to a pool of parent solutions which are then combined with fixed-point crossover.

\section{Experimental Study}
\label{sec:Exp}
The following section outlines the experimental data-streams used to validate the proposed method, along with the metrics used. A comparison of both bio-inspired population based approaches is presented along with a comparison with peer algorithms. A complexity analysis is offered in Section \ref{sec:CA}. 

\subsection{Data Streams}
\label{sec:data}
Ten dynamic data streams are used to validate the proposed methods. These are taken from the non-stationary archive page\footnote{https://sites.google.com/site/nonstationaryarchive/} and have been used in previous studies on non-stationary data streams \cite{COCEL, scarg} and also in the EVL setting \cite{amanda, slayer}. 

The characteristics of each stream is presented in Table \ref{tab:data} (and a visualisation of the streams can be seen from the non-stationary archive page, footnote 1) . The number of classes ranges from 2 to 4, and the dimensionality ranges from 2 to 10. Although some of the streams appear simple, they are non trivial. For example, synthetic stream 4CR consists of four rotating classes. Each class gradually moves into an area of the feature space previously occupied by a different class. So, along with adaptation, some form of `forgetting' is required. Most are synthetic streams and a real stream is included in Keystroke data. This stream is based on keystroke dynamics, users are recognised by their typing rhythm which is likely to change over time as concept drift. The stream describes 4 different users typing a 10-key password 400 times. The 10 variables measure the time difference between a key being released and the next one being pressed.
\begin{table}[thb]
\centering
\caption{Comparison with Peer EVL methods}
\label{tab:compPeer}
\resizebox{\columnwidth}{!}{
\begin{tabular}{ccccc}
\toprule
 	Dataset & COMPOSE & LEVELiw & PSO & GA \\
		\midrule
		$1CDT$  & \textbf{0.99} & \textbf{0.99} & \textbf{0.99(.002)} & \textbf{0.99(.002)(=)}  \\
		$4CR$ & \textbf{0.99} & \textbf{0.99} & \textbf{0.99(.001)} & \textbf{0.99(.001)(=)} \\
		$1CSurr$ & 0.90 & 0.63 & \textbf{0.95(.004)} &  \textbf{0.95(.024)}(+)  \\
		$GEARS$ & \textbf{0.96} & 0.93  & 0.82(.001) & 0.91(.003)(-)  \\
		$UG\_2C\_2D$  & 0.94 & 0.73 & 0.91(.000) & \textbf{0.96(.001)}(+)   \\
		$UG\_2C\_3D$ & 0.67 & 0.64 & 0.96(.001) & \textbf{0.97(.001)}(+) \\
		$UG\_2C\_5D$ & 0.90 & 0.60  & 0.86(.001) & \textbf{0.92(.03}(+)   \\
		$4CRE\_V1$ & 0.20 & 0.24 & 0.28(.031) &  \textbf{0.32(.061)}(+)  \\
		$4CRE\_V2$ & 0.19 & 0.24 & 0.25(.061) &  \textbf{0.33(.067)}(+) \\
		$Keystroke$ & \textbf{0.85} & 0.78 & 0.73(.023) & 0.76(.004) (-) \\
\midrule
\bottomrule
\end{tabular}
}
\end{table}
\subsection{Metrics}
\label{sec:metric}
The performance of the proposed method is measured using the F1-Score (sometimes called the F-Score or F1-measure). It is the mean of the precision and recall scores of the algorithm over a given window $w_i$. We use the macro-F1 score as it is suitable to use for balanced and unbalanced data-sets (the alternative micro-F1 score favours the majority class). The final result is the average of all of the windows. An incoming point $x_i$ with a true label $y_i$ is passed to the ensemble E and a prediction $\hat{y}_i$ is returned; $\hat{y}_i = E(x_i)$.

The predicted values are compared with the actual values and precision and recall scores calculated. Combining these metrics gives the F1-Score:
\begin{equation}
F1 = 2 * \frac{Precision * Recall}{Precision + Recall}
\label{eq:f1}
\end{equation}

When evaluating the population-based algorithms, we run each experiment 20 times. To evaluate whether the resulting distributions are significantly different, we use the Wilcoxen Signed-Rank test. We reject the null hypothesis that two distributions are equal with $p < 0.05$. When comparing the population algorithm results with peer results published in the literature we use the same statistical test and reject the null hypothesis that the distribution of results are symmetric around the peer results with $p < 0.05$

\subsection{Evaluation}
\label{sec:pop_comp}
We begin by evaluating the datastreams on a $static$ ensemble. In these experiments the ensemble is initialised with a labelled training set but is never updated as the stream progresses. In these instances the ensemble begins with a high accuracy but does not react to change.
As expected, the ensemble performs poorly over the course of the non-stationary stream. This is because the constituent classifiers remain specialised in areas of the feature space which are no longer relevant to the underlying distribution. Virtual concept drift or real drift has occurred and the ensemble has not reacted resulting in misclassified instances. 

\begin{figure*}[!thb]
\centering
\includegraphics[scale=0.45]{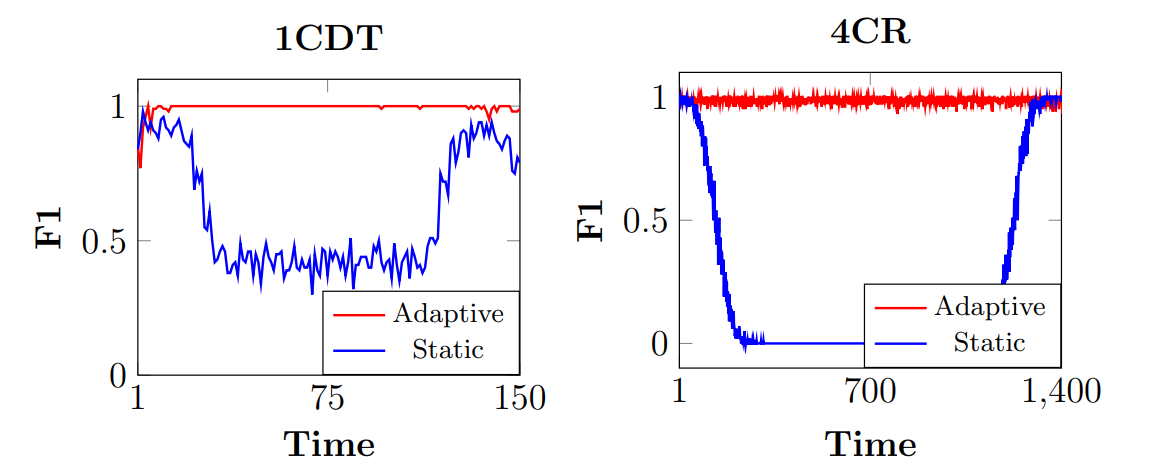}
\caption{Static Ensemble vs adaptive ensemble}
\label{fig:staticComp}
\end{figure*}
\begin{figure*}[!thb]
\centering
\begin{multicols}{5}[\columnsep=0.18cm]
\hspace{-.5cm}
\includegraphics[scale=0.18]{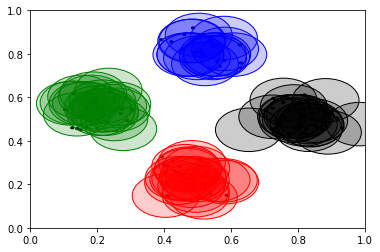}
\caption*{t = 0}
\columnbreak
\includegraphics[scale=0.18]{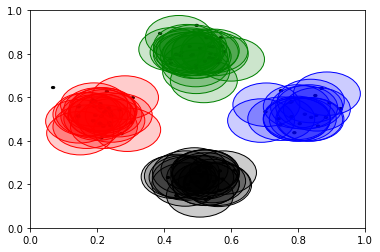}
\caption*{t = 350}
\columnbreak
\includegraphics[scale=0.18]{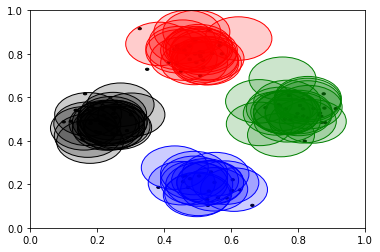}
\caption*{t = 700}
\columnbreak
\includegraphics[scale=0.18]{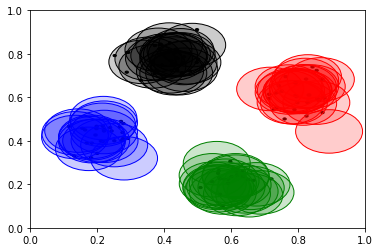}
\caption*{t = 1050}
\columnbreak
\includegraphics[scale=0.18]{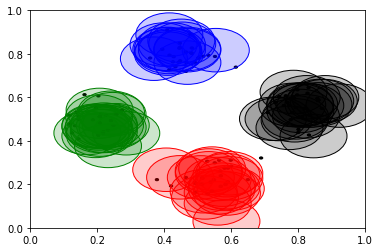}
\caption*{t = 1400}
\end{multicols}
\caption{Stream 4CR, 4 classes rotate. Ensemble adapts to underlying change using Genetic Algorithm to update population of classifiers}
\label{fig:4CRprogress}
\end{figure*}

As an illustrative example, the performance of the static ensemble is presented alongside the performance of the adaptive ensemble in Fig. \ref{fig:staticComp}. Here, a GA is used to adapt to the underlying change. The left plot shows the progression on the 1CDT stream. Here, two classes are present. One is stationary and the other dynamic. The static ensemble maintains $\approx 50\%$  accuracy because of the stationary class but does not track the non-stationary class. The GA does and maintains a high accuracy through the stream. In the right plot, stream 4CR is used. This stream consists of four classes which rotate and move into positions previously occupied by other classes. The static ensemble starts with high accuracy but as the stream progresses the accuracy drops. It rises again once the classes complete their rotation and return to the starting position. In Fig. \ref{fig:4CRprogress}, this process is visualised; the adaptation of the ensemble with a Genetic Algorithm is presented. The drift of the 4 classes can be observed as the algorithm adapts to this drift.

\subsection{Performance Comparison with Peer Methods}
\label{sec:peer_comp}
In this subsection, we tabulate the results of the ensembles using a GA and PSO adaptation method and compare their performance with other EVL methods. We evaluate the proposed method with COMPOSE \cite{compose} and LEVELiw \cite{levelIW}. The peer-results are taken from results previously published in the literature. The results are presented in Table \ref{tab:compPeer}. The proposed approach is favourable to both peer algorithms.

\subsection{Complexity Analysis}
\label{sec:CA}
The memory requirements of the proposed method is $O(NC)$ where $N$ is the number of agents per class and $C$ is the number of classes in the stream (assuming this number does not increase during the stream). The time complexity of the solution is similar. In the classification phase, the time requirement is also $O(NC)$ and the population update requires a further $O(NC)$ as this depends on the number of ensembles and agents per ensemble. This gives a total time complexity of $O(NC^2)$.


Of both methods evaluated, PSO is the fastest because it requires the least amount of computation to update the population. However, it returns the lowest accuracy. Conversely, the GA is slower but returns the highest accuracy. The time requirement is a function of the number of agents, the more agents the more calculations and the longer the algorithm takes to process the stream. As an illustrative example the time requirement for two streams is presented. In Table \ref{tab:1CDTagents}, 1CDT stream is presented and the number of seconds along with the F1 score is presented as the number of agents in each ensemble increases. PSO is faster than the GA with the same F1 accuracy and increasing the number of agents does not improve accuracy despite taking longer to process. In Table \ref{tab:5Dagents}, it can be seen that PSO is again much faster but achieves a lower F1 score. The GA takes more time but gives a better accuracy and this accuracy increases with a larger number of agents.

\begin{table}[!bth]
\centering
\caption{Number of agents per class in 1CDT Stream}
\label{tab:1CDTagents}
	\begin{tabular}{lrrcrr}\toprule
		& \multicolumn{2}{c}{PSO} & \phantom{ab}& \multicolumn{2}{c}{GA} \\
		\cmidrule{2-3} \cmidrule{5-6} 
		Agents & F1 & Secs. && F1 & Secs.  \\ \midrule
		10 & 0.98& 0.3 && 0.99 & 0.4 \\
		20 & 0.99& 0.57 && 0.99 & 0.8 \\
		50 & 0.99& 1.12 && 0.99 & 2.0\\
		100 & 0.99& 2.4 && 0.99 & 4.5 \\		
		\bottomrule		
	\end{tabular}%
\end{table}

 \begin{table}[!bth]
\centering
\caption{Number of agents per class in UG\_2C\_5D Stream}
\label{tab:5Dagents}
		\begin{tabular}{lrrcrr}\toprule
		& \multicolumn{2}{c}{PSO} & \phantom{ab}& \multicolumn{2}{c}{GA}  \\
		\cmidrule{2-3} \cmidrule{5-6} 
		Agents & F1 & Secs. && F1 & Secs. \\ \midrule
		10 & 0.81& 5.5 && 0.81 & 7.0 \\
		20 & 0.84& 10.3 && 0.85 & 12.6 \\
		50 & 0.86& 19.6 && 0.91 & 31.7 \\
		100 & 0.86& 46.1 && 0.92 & 73.0 \\
		\bottomrule		
	\end{tabular}%
\end{table}

\section{Conclusions and Future Work}
\label{sec:conc}
In this paper we proposed an evolving population-based approach to the problem of classification in data-streams with extreme verification latency (EVL). Two population based approaches were examined; PSO and GA with the GA giving the best results. These algorithms were evaluated on a range of data-streams and the population based approach shows promise when compared to peer EVL algorithms. An ensemble of one-class classifiers is maintained for each class whereby each classifier is considered an agent in the sub-population. After each window the population is subject to selection pressure and in this way the ensemble reacts to underlying change in the stream and maintains high classification rate throughout the dynamic stream without access to any further labels beyond the initial training set.

The research serves as an initial study in the use of evolutionary algorithms for addressing the problem of non-stationary and EVL in data-streams. It also reveals a number of future directions. In this study, each ensemble is composed of relatively simple base-classifiers (a micro-classifier, similar to $k$-NN). Performance could be improved with more sophisticated one-class classifiers such as Minimum Spanning Trees \cite{MST}, One Class Support Vector Machine \cite{SVMOCC}, or others \cite{OCCsurvey}, or perhaps a heterogeneous combinations thereof (for example more heavy-weight sophisticated classifiers in stable areas of the feature space with simpler lightweight models which can be deployed quickly and provide a better response to change). There is also potential for improving/specialising the optimisation techniques for updating the ensemble. For example, PSO requires fewer calculations but does not provide as accurate performance. These are considerations which might be stream dependant and this is an area for future research. The study makes some assumptions which are common in the EVL literature. Firstly, it is assumed that the data-stream will not exhibit concept evolution, so classes which are not present in the initial training data will not appear later in the stream. Secondly, change in the stream is assumed to be gradual. The proposed method can capture real drift assuming that the change in $P(y|x)$ begins as virtual drift in $P(x)$. However, sudden real drift (a change in $P(y|x)$ without change in $P(x)$) is difficult to recognise without access to the true class label (or a human expert) and is an important future research direction.

\vspace{12pt}

\end{document}